\documentclass[runningheads]{llncs}
\usepackage{graphicx}
\usepackage{placeins}
\usepackage{bbding}
\usepackage{amsfonts}
\graphicspath{{Images/}}
\begin{document}
\title{Perceptual Embedding Consistency for Seamless Reconstruction of Tilewise Style Transfer}
\titlerunning{Perceptual Embedding Consistency for Seamless Reconstruction}
%

\author{Amal Lahiani\inst{1,2} $^{\textrm{\Envelope}}$ \and
Nassir Navab\inst{2} \and
Shadi Albarqouni\inst{\thanks{Shared senior authorship.} \: 2} \and
Eldad Klaiman \inst{\footnotemark[1]  \: 1}}
\authorrunning{Lahiani et al.}
\institute{Pathology and Tissue Analytics, Pharma Research and Early Development, Roche Innovation Center Munich \\
\email{amal.lahiani@roche.com} \and
Computer Aided Medical Procedures, Technische Universit\"at M\"unchen \\
}


%
\maketitle              
\begin{abstract}
Style transfer is a field with growing interest and use cases in deep learning. Recent work has shown Generative Adversarial Networks (GANs) can be used to create realistic images of virtually stained slide images in digital pathology with clinically validated interpretability. Digital pathology images are typically of extremely high resolution, making tilewise analysis necessary for deep learning applications. It has been shown that image generators with instance normalization can cause a tiling artifact when a large image is reconstructed from the tilewise analysis. We introduce a novel perceptual embedding consistency loss significantly reducing the tiling artifact created in the reconstructed whole slide image (WSI). We validate our results by comparing virtually stained slide images with consecutive real stained tissue slide images. We also demonstrate that our model is more robust to contrast, color and brightness perturbations by running comparative sensitivity analysis tests.

\end{abstract}
\section{Introduction}

In the field of pathology tissue staining is used to examine biological structures in tissue. Tissue staining is a complex, expensive and time consuming process. Additionally, tissue samples are scarce and expensive. As a result, different state of the art style transfer based methods have been applied in order to synthesize virtually stained images from other modalities. Style transfer is a field with growing interest and use cases in deep learning allowing to render an image in a new style while preserving its original semantic content. One of the main challenges of style transfer applications is the necessity to distinguish between style features (e.g. color) and content features (e.g. semantic structures) \cite{1}. Recent deep learning based style transfer works have shown that using perceptual losses instead of or along with pixel level losses can help the network learn relevant high level style and content features and thus generate high quality stylized images \cite{7,8}. Deep learning based style transfer has been used to generate augmented faces \cite{2}, virtual artwork with specific artist styles \cite{3} and recently also virtually stained histopathological whole slide images (WSIs) \cite{5,6}.

Some groups used approximative and empirical methods in order to virtually generate H\&E images from fluorescence images \cite{12,13}. In the field of deep learning, generative Adversarial Networks (GANs) have been used in \cite{15,16,17} in order to predict brightfield images from autofluorescence of unstained tissue, H\&E from unstained lung hyperspectral tissue image and immunofluorescence from H\&E respectively. In \cite{6} and \cite{19}, a neural network has been used in order to predict fluorescent labels from transmitted light images. The training of these supervised methods is based on spatially registered image pairs of the input and output modalities. As generating paired slide images with different stainings is a complex task involving the use of consecutive tissue sections or a stain-wash-stain technique, unsupervised deep learning methods have been used in virtual staining \cite{5} and stain normalization applications \cite{22}. In \cite{5}, CycleGAN \cite{3} has been used in order to virtually generate duplex Immunohistochemistry (IHC) stained images from real stained images. 

Another important challenge in digital pathology computer aided applications is the size of high resolution WSIs. For this reason, tile based analysis is usually used in order to deal with memory limitations. Inference of trained generators on WSIs in a tilewise manner when instance normalization modules are used has been shown to present tiling artifacts in the reconstructed WSI \cite{5}. While instance normalization has been proven to be crucial in style transfer applications GAN training \cite{21}, it makes a pixel in the output image depend not only on the network and the receptive field area but also on the statistics of the entire input image. This results in applying different functions to adjacent pixels belonging to different adjacent tiles. Using large overlap for the tiles during inference can mitigate this problem \cite{5} but still presents some residual tiling and is very costly.

In order to make virtual staining of WSIs more efficient and robust for real world use cases, we aim to better address this instance normalization induced tiling artifact in a way that does not require superfluous processing. We introduce a novel and new perceptual embedding loss function into the CycleGAN network architecture during training aimed at regularizing the effect of input image contrast, color and brightness perturbations in the generator latent space. We apply our proposed method to train a network to generate a virtual brightfield IHC fibroblast activation protein (FAP) - cytokeratin (CK) duplex tissue staining images from stained H\&E tissue images. We validate our results by comparing the virtually generated images to their corresponding real consecutive stained slides. We also perform a comparative sensitivity analysis to validate our hypothesis that the introduced perceptual embedding loss helps train a generator network that is more contrast, brightness and color perturbation robust.

\section{Proposed Approach}
\begin{figure}[t]

  \centering
  \centerline{\includegraphics[width=0.9\linewidth]{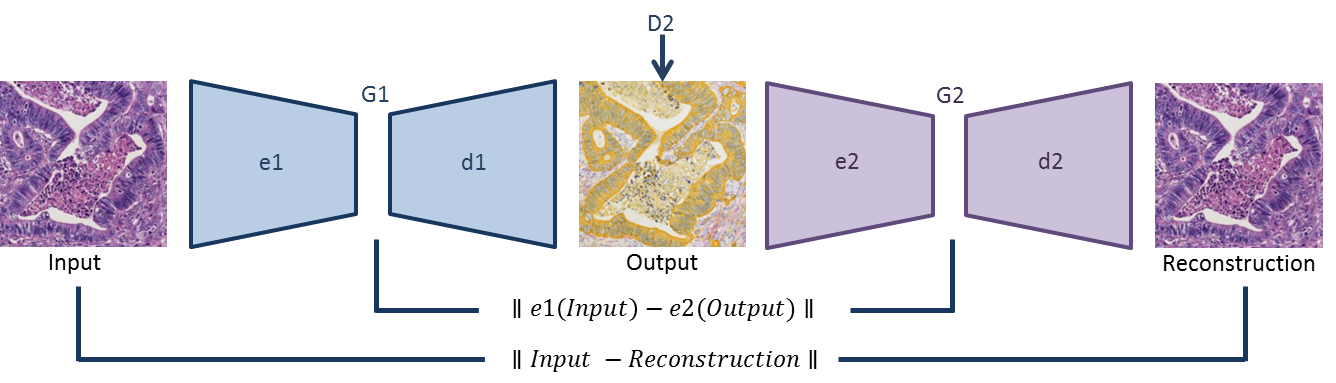}}
  
\caption{Proposed model. G, D, e and d denote generator, discriminator, encoder and decoder respectively. The objective function includes three loss types: adversarial loss, cycle consistency loss and color invariant embedding consistency loss.
}
\label{peliGAN}
\end{figure}

We propose a novel approach to generate seamless high quality virtual staining WSIs and specifically address the image reconstruction tiling artifact by introducing a perceptual embedding consistency loss to the CycleGAN network during training. CycleGAN model is built under the assumption that translating one image from domain X to Y then back from Y to X should result in the original input image. It consists of two mapping generators and two discriminators aiming at distinguishing between the real and generated images. The addition of the perceptual embedding loss allows to minimize the difference between the latent features in the two generators of the CycleGAN (Fig. \ref{peliGAN}). We use L2-norm to calculate the distance between the latent features in the generators bottlenecks and add this loss multiplied by a weight to the total loss of the network architecture alongside the reconstruction and adversarial losses. 

The combination of these different losses helps the network learn meaningful feature-level representations in an unsupervised fashion in order to capture the semantics and styles of input and output histology staining domains, thus allowing us to learn a meaningful mapping between both domains and to obtain a more homogeneous contrast in virtual WSIs after tilewise inference. We hypothesize that forcing bottleneck features in both generators to be similar forces color and contrast invariance. Color and contrast invariant features that could successfully enable generation of virtual histopathological staining while maintaining the capability to cycle back to the original image could consist of condensed anatomical information, e.g. cell shapes, nuclear density, tissue textures, etc. Under this assumption, adjacent pixels belonging to different tiles would be more homogeneously mapped to the output space.

The objective function for our network includes three loss types. In addition to the adversarial loss and the cycle consistency loss described in \cite{3} we add a perceptual embedding consistency loss (equation \ref{L_embd}) between the two generator embeddings. This perceptual loss forces the generators in the network to learn a semantic content and contrast free features in the latent space, allowing a homogenization of the output contrast when the new style is added to the semantic features in the decoder block. We introduce the embedding consistency loss as:

\begin{equation}
L_{embd}(G,F) = \mathbb{E}_{X \sim \mathbb{P}_{X}}[\| e_1(x) - e_2(G_1(x))\|_{2}] + \mathbb{E}_{Y \sim \mathbb{P}_{Y}}[\| e_2(y) - e_1(G_2(y))\|_{2}],
\label{L_embd}
\end{equation}
where $X$ and $Y$ correspond to the two domains. $G_1$ and $G_2$ correspond to the generators of the model, $e_1$ and $e_2$ correspond to the encoders of the first and second generator respectively and $\|.\|_{2}$ is the $L2$ distance. The combined objective function is then:

\begin{equation}
L = L_{GAN}(G_1,D_2,X,Y)+L_{GAN}(G_2,D_1,Y,X)+\omega _{cyc} L_{cyc}+\omega _{embd} L_{embd},
\label{L_total}
\end{equation}
where $D_1$ and $D_2$ correspond to both discriminators, $L_{GAN}(G_1,D_2,X,Y)$ and $L_{GAN}(G_2,D_1,Y,X)$ to the adversarial losses of both mappings and $L_{cyc}$ to the cycle consistency loss. $\omega _{cyc}$ and $\omega _{embd}$ correspond to the weights of the cycle and embedding consistency losses respectively.

\section{Experiments and Results}
We use the described approach to train a network to virtually stain biopsy tissue WSIs with a duplex FAP-CK IHC stain from H\&E stained slide images. CK is a marker for tumor cells and (FAP) is expressed by cancer-associated fibroblasts in the stroma of solid tumor. H\&E is widely used for cancer diagnosis and tissue assessment, so our application could leverage existing H\&E images to generate new and otherwise unattainable information about these biopsies.

\subsection{Dataset}
Our dataset consists of a selected set of WSIs from surgical specimen of Colorectal Carcinoma metastases in liver tissue from our internal pathology image database. The dataset includes 25 tissue blocks from different patients, each with 2 consecutive slides stained with H\&E and FAP-CK respectively. We divide the total of the 50 WSIs into training and testing sets with 10 WSIs and 40 WSIs respectively.
Our training set consists of 5 H\&E stained WSIs and 5 FAP-CK stained WSIs from 5 patients tissue blocks. Due to memory limitations, all high resolution training images were split into $512 \times 512$ tiles with 128 overlap at 10x magnification factor. After tiling, our training dataset contains 7592 H\&E $512 \times 512$ RGB tiles and 7550 FAP-CK $512 \times 512$ RGB tiles.
We validate our method on a dedicated test dataset consisting of 20 paired WSIs from consecutive tissue sections of the same tissue blocks stained with H\&E and FAP-CK . The testing images are taken from different patients than those of the training set.

\subsection{Implementation Details}
We trained the proposed model and a baseline CycleGAN model \cite{5} for 20 epochs with ResNet-6 generator architectures and 70x70 PatchGANs discriminators. Hyperparameters are chosen similarly to \cite{3} and we fix the embedding weight $\omega _{embd}$ to be equal to the cycle consistency loss weight $\omega _{cyc}$ ($\omega _{embd}=\omega _{cyc}=10$). The training was distributed on multiple GPUs using the Pytorch distributed computing library and the stochastic synchronous ADAM algorithm. We train the model on a High Performance Computing (HPC) Cluster.

\subsection{Evaluation Metric}
We evaluate and compare our method to the CycleGAN architecture described in \cite{5} by measuring the complex wavelet structural similarity index CWSSIM \cite{20} between the virtually generated FAP-CK images and their corresponding consecutive real stained FAP-CK images. In order to use the CWSSIM as an evaluation metric, we registered the consecutive slide images using a geometric point set matching method. 

CWSSIM is an extension of the structural similarity index (SSIM) to the complex wavelet domain. The CWSSIM index is a number bounded between 0 and 1 where 0 means highly dissimilar images and 1 means perfect matching. The choice of the CWSSIM as a measure to assess the performance is based on the fact that this metric, unlike other metrics, is robust to small translations and rotations. Actually, as explained in \cite{20}, small image distortions result in consistent phase changes in the local wavelet coefficients which does not affect the structural content of the image. Since our validation paired real/virtual data are not obtained from the same section but from consecutive sections of the same tissue block, even after careful registration of the paired images, mismatched areas in the tissue could still exist in some regions of the images. This fact made the use of other intensity and geometric based indices not necessarily very well correlated to the visual similarity between the images.

\subsection{Ablation Test and Comparison}
\begin{figure}[t]

\begin{minipage}[b]{.24\linewidth}
  \centering
  \centerline{\includegraphics[width=0.9\linewidth]{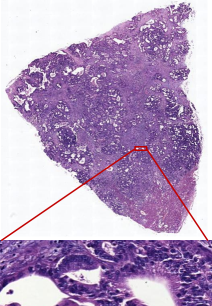}}
  \centerline{(a)}\medskip
\end{minipage}
\hfill
\begin{minipage}[b]{.24\linewidth}
  \centering
  \centerline{\includegraphics[width=0.9\linewidth]{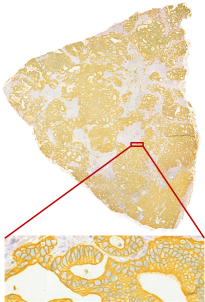}}
  \centerline{(b)}\medskip
\end{minipage}
\hfill
\begin{minipage}[b]{.24\linewidth}
  \centering
  \centerline{\includegraphics[width=0.9\linewidth]{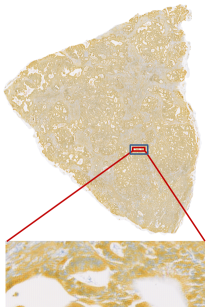}}
  \centerline{(c)}\medskip
\end{minipage}
\hfill
\begin{minipage}[b]{.24\linewidth}
  \centering
  \centerline{\includegraphics[width=0.9\linewidth]{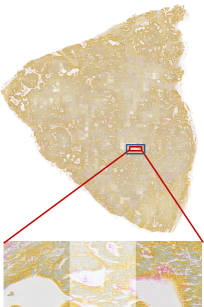}}
  \centerline{(d)}\medskip
\end{minipage}
\begin{minipage}[b]{.45\linewidth}
  \centering
  \centerline{\includegraphics[width=0.7\linewidth]{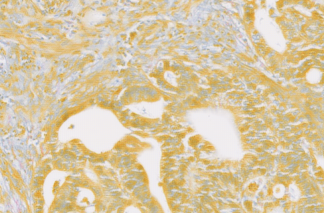}}
  \centerline{(e)}\medskip
\end{minipage}
\hfill
\begin{minipage}[b]{.45\linewidth}
  \centering
  \centerline{\includegraphics[width=0.7\linewidth]{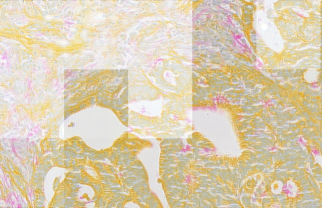}}
  \centerline{(f)}\medskip
\end{minipage}
%
\caption{(a), (b), (c) and (d) correspond to an input H\&E image from the testing set, the corresponding consecutive registered real stained FAP-CK, the virtual FAP-CK image obtained with our model and the virtual FAP-CK image obtained with the baseline CycleGAN. (e) and (f) correspond to zooms of the blue boxes in (c) and (d) respectively. We can clearly see the effect of the embedding consistency loss in homogenising the contrast of the reconstructed WSI.}
\label{embd_plain}
\end{figure}

In order to visualize the effect of our proposed perceptual embedding consistency loss, we conduct an ablation study where we train the same model using the same generators and discriminators architectures, the same hyperparameters and the same number of epochs with and without the perceptual embedding consistency loss. 
We notice that both models learn a reasonable mapping between domains in the field of view level and the semantic content of the images is generally preserved. However, when we consider the reconstructed WSIs, we notice that our approach yields a significantly more continuous image with substantially less tiling artifacts than the baseline method (Fig. \ref{embd_plain}).

Virtual and consecutive real images are first registered and tiled to $1024 \times 1024$ fields of views. The 20 testing blocks yielded 1236 virtual tiles and their corresponding 1236 real tiles. Then we compute the similarity index between the real and virtual tiles for the models trained with and without the embedding consistency loss. The results are summarized in Table~\ref{tab1}. The median CWSSIM for all patients is equal to 0.79 and 0.74 with our approach and CycleGAN respectively, reflecting 6.75\% of relative improvement. Additionally, we measure the CWSSIM per patient and we observe higher average CWSSIM for 85\% of the patients. 

\begin{table}
\caption {Results of CWSSIM index between real and virtual samples.}\label{tab1}
\centering
\setlength{\tabcolsep}{10pt}
\begin{tabular}{c|c}
\hline
Method & Mean (Median) $\pm$ Std\\
\hline
CycleGAN \cite{5} & 0.74 (0.74) $\pm$ 0.153\\
Ours & \textbf{0.77} (\textbf{0.79}) $\pm$ \textbf{0.146}\\
\hline
\end{tabular}
\end{table}

\subsection{Sensitivity Analysis}
In order to verify our assumptions about the effect of the perceptual embedding consistency loss on learning semantic content and a more color, contrast and brightness invariant embedding, we perform a comparative sensitivity analysis. The analysis includes insertion of color, brightness and color perturbations into the generator input and comparison of the effect of that perturbation on the generator embeddings between our approach and plain CycleGAN. 

\begin{figure}[t]

\begin{minipage}[b]{.3\linewidth}
  \centering
  \centerline{\includegraphics[width=1.0\linewidth]{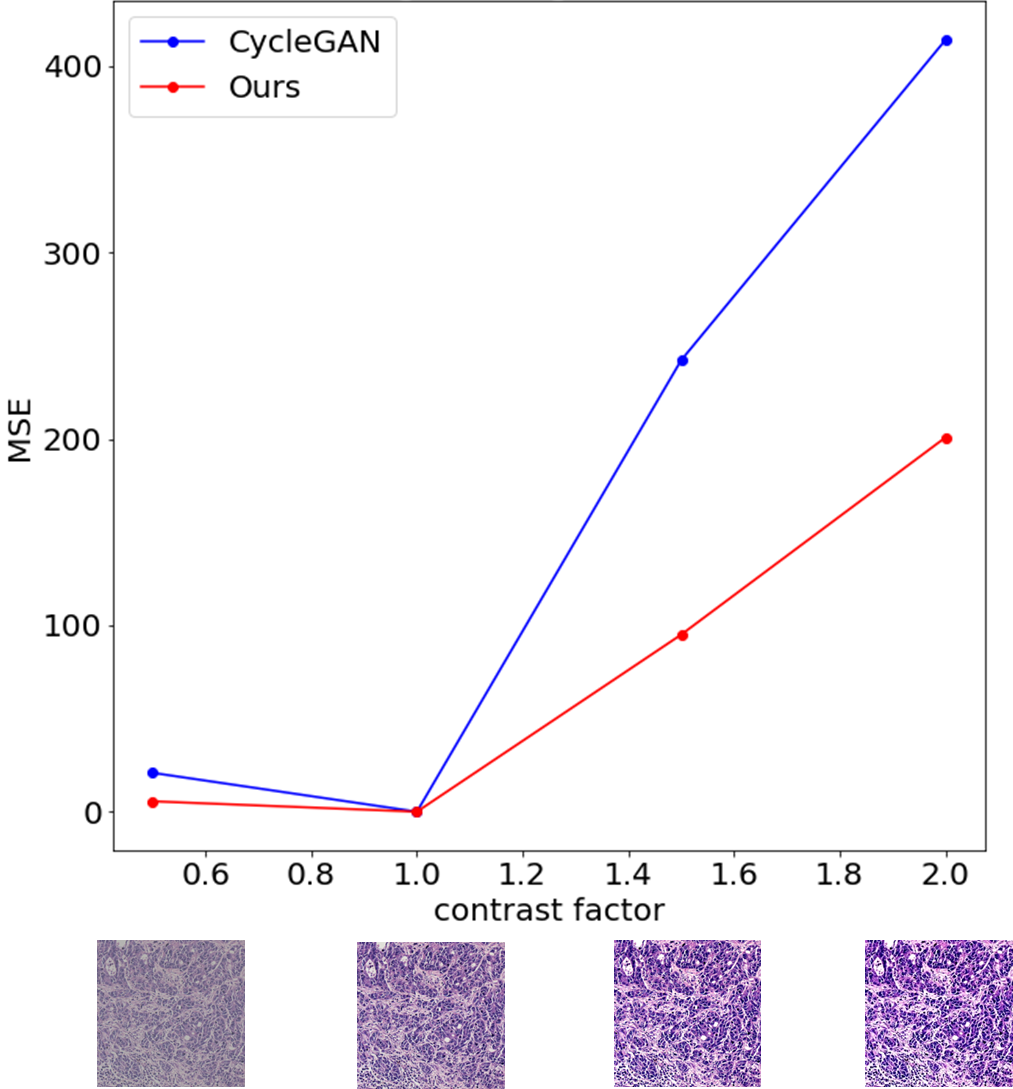}}
  \centerline{(a)}\medskip
\end{minipage}
\hfill
\begin{minipage}[b]{0.3\linewidth}
  \centering
  \centerline{\includegraphics[width=1.0\linewidth]{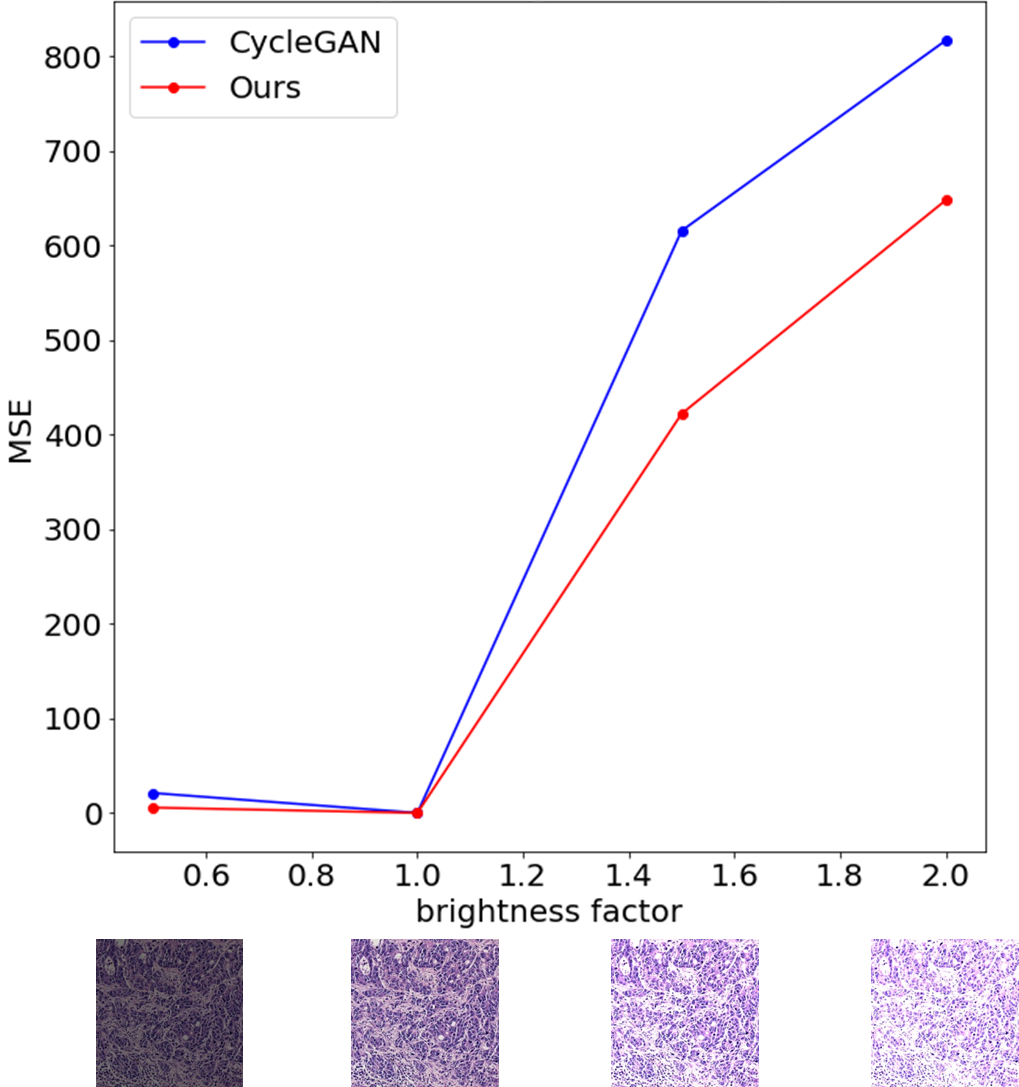}}
  \centerline{(b)}\medskip
\end{minipage}
\hfill
\begin{minipage}[b]{0.3\linewidth}
  \centering
  \centerline{\includegraphics[width=1.0\linewidth]{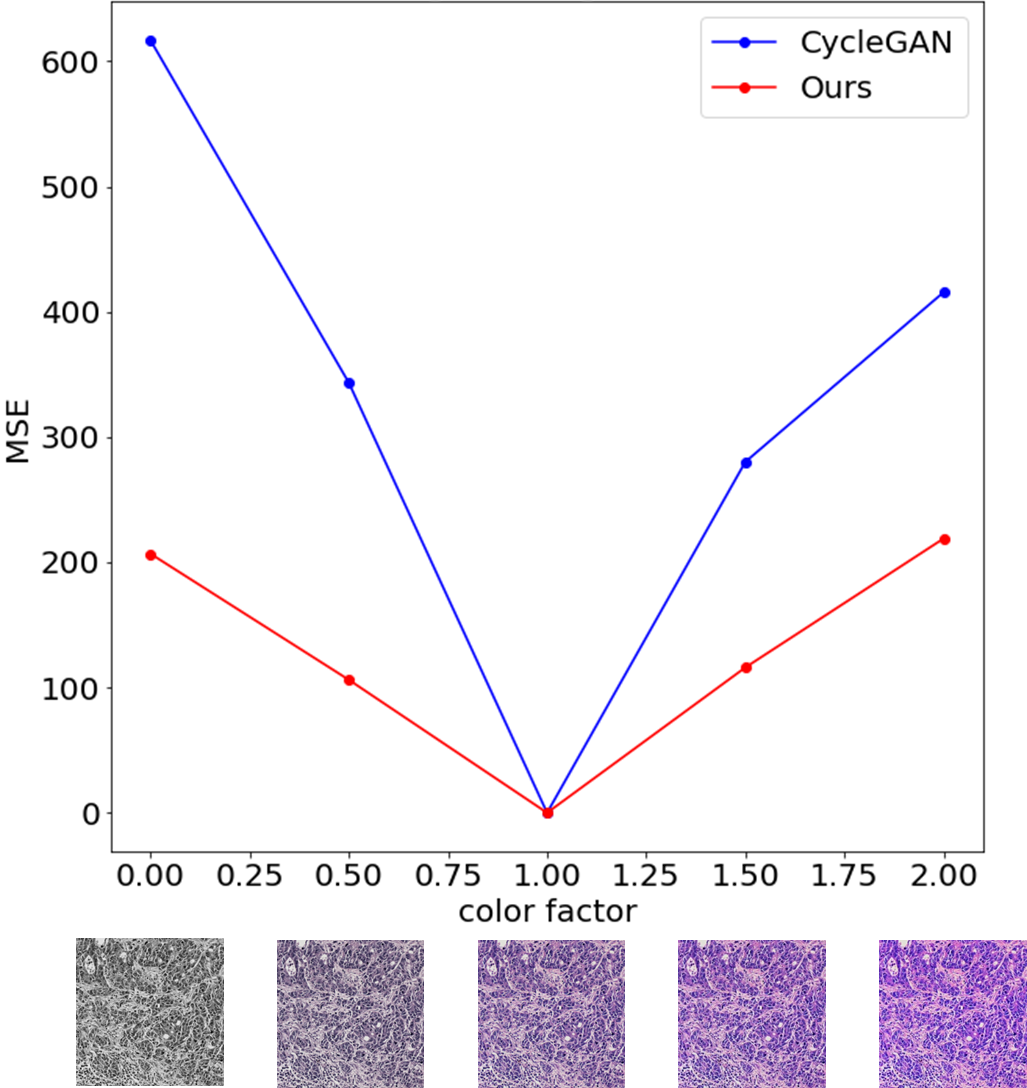}}
  \centerline{(c)}\medskip
\end{minipage}
%
\caption{Average MSE between the embeddings of the original and perturbed tiles for the 100 selected tiles. (a), (b) and (c) correspond to contrast, brightness and color perturbations respectively. The blue and red curves correspond to the results obtained for CycleGAN and our approach respectively.}
\label{sensitivity}
\end{figure}

For this, we randomly select a subset of 100 ($512 \times 512$) tiles from the testing dataset. We perform an inference with the model’s generator on each tile followed by inference of the same generator on different perturbed versions of the tile. Then we calculate the Mean Square Error (MSE) between the embeddings of the original tiles and embeddings of each of the corresponding perturbed versions of the tiles. 
Fig. \ref{sensitivity} shows the average MSE values of the 100 selected tiles obtained from contrast, brightness and color perturbations of the generator input. We report on the results obtained with our approach compared with the baseline CycleGAN. All graphs clearly show that our approach results in smaller MSE values in the latent space for the different perturbations. This shows that the perceptual embedding loss drives the network to learn image embeddings that are more content related and more invariant to color, brightness and contrast changes. These robust invariant embeddings enable the network to be more robust to the effects of global changes in tile statistics and results in smoother seamless WSI reconstruction despite the effects of the instance normalization module. 

\section{Discussion and Conclusion}

We present a novel style transfer approach applied to the field of digital pathology where high resolution WSIs are required. In this specific application, we virtually generate FAP-CK images from real stained H\&E images using tilewise processing because of hardware memory limitations. In particular, we propose a solution based on perceptual embedding consistency loss in order to obtain a substantially more homogeneous contrast in the WSIs. 

We demonstrate that this targeted regularization forces the network to learn non-color and non-contrast based features for tile embedding and this in turn reduces variation of the output tiling artifact due to the instance normalization effect. While the proposed solution seems to improve the results and to solve one of the main issues we had to deal with in our stain virtualization frameworks, there is still a lot to investigate and a lot of room for improvement. 
For example, we plan to study the effect of the input staining biological features on the quality of the virtual images. Additionally, we noticed that, unlike CK which was very well correlated in real and virtual images, FAP reconstruction is still showing significant differences. We plan to investigate the problem to understand if it is correlated to biological constraints (i.e. lack of predictive features in the input staining) or to weaknesses in the architecture. For this reason we plan to add more constraints to the network in order to investigate the challenge and, when possible, improve the quality of FAP reconstruction.

%
%
%
\FloatBarrier
\bibliographystyle{splncs04}
\bibliography{references}
\end{document}